\def\year{2016}\relax
\documentclass[letterpaper]{article}
\usepackage{aaai17}
\usepackage{url}
\usepackage{graphicx}
\usepackage{epstopdf}
\usepackage{amsfonts}
\usepackage{amssymb}
\usepackage{amsmath}
\usepackage{verbatim}
\usepackage{multirow}
\usepackage{subfigure}
\usepackage{array}
\usepackage{fp}
\usepackage{makecell}
\usepackage{flushend}
\usepackage{cases}
\usepackage{color}
\usepackage{times}
\usepackage{url}
\usepackage{latexsym}
\usepackage{times}
\usepackage{url}
\usepackage{graphicx}
\usepackage{epstopdf}
\usepackage{amsfonts}
\usepackage{amssymb}
\usepackage{amsmath}
\usepackage{verbatim}
\usepackage{multirow}
\usepackage{subfigure}
\usepackage{array}
\usepackage{fp}
\usepackage{makecell}
\usepackage{flushend}
\usepackage{cases}
\usepackage{color}
\usepackage{bbm}
\usepackage{times}
\usepackage{helvet}
\usepackage{courier}
\newcommand{\thickhline}{\noalign{\hrule height 1pt}}
\begin{document}
	%
	\title{Response Selection with Topic Clues for Retrieval-based Chatbots}
	\author{
		Yu Wu$^\dag$\thanks{ The work was done when the first author was an intern in Microsoft Research Asia.}~~~~, Wei Wu$^\ddag$~~~~, Zhoujun Li$^\dag$~~~~, Ming Zhou$^\ddag$~~~~\\
		$^\dag$State Key Lab of Software Development Environment, Beihang University, Beijing, China\\
		$^\ddag$~~~~Microsoft Research, Beijing, China\\
		\{wuyu,lizj\}@buaa.edu.cn \{wuwei,mingzhou\}@microsoft.com 
	}
	\maketitle
	\begin{abstract}
		We consider incorporating topic information into message-response matching to boost responses with rich content in retrieval-based chatbots. To this end, we propose a topic-aware convolutional neural tensor network (TACNTN). In TACNTN, matching between a message and a response is not only conducted between a message vector and a response vector generated by convolutional neural networks, but also leverages extra topic information encoded in two topic vectors. The two topic vectors are linear combinations of topic words of the message and the response respectively, where the topic words are obtained from a pre-trained LDA model and their weights are determined by themselves as well as the message vector and the response vector. The message vector, the response vector, and the two topic vectors are fed to neural tensors to calculate a matching score. Empirical study on a public data set and a human annotated data set shows that  TACNTN can significantly outperform state-of-the-art methods for message-response matching.
	\end{abstract}
	
	\section{Introduction}
	Human-computer conversation is a challenging task in AI and NLP. Existing conversation systems include task oriented dialog systems and non task oriented chatbots. The former aims to help people complete specific tasks such as ordering and tutoring, while the latter focuses on talking like a human and engaging in social conversations regarding a wide range of issues within open domains \cite{perez2011conversational}. Although previous research on conversation focused on dialog systems, recently, with the large amount of conversation data available on the Internet, chatbots are drawing more and more attention in both academia and industry.  
	The key problem for building a chatbot is how to reply to a message with a proper (human-like and natural) response. Existing methods are either retrieval-based or generation-based. Retrieval-based methods \cite{ji2014information} retrieve response candidates from a pre-built index, rank the candidates, and select a reply from the top ranked ones, while generation-based methods \cite{DBLP:conf/acl/ShangLL15,vinyals2015neural} leverage natural language generation (NLG) techniques to respond to a message. In this work, we study response selection for retrieval-based chatbots in a single turn scenario, because retrieval-based methods can always return fluent responses \cite{ji2014information} and single turn is the basis of conversation in a chatbot.

	\begin{table}
		\caption{A good response to a message}	
		\centering
		\begin{tabular}{m{8cm}}
			\hline
			\textbf{Message} : Is the new Batman movie worth watching?  \\ \hline			
			\textbf{Response} : I swear you won't regret watching it. We finally get Batman as a fully rendered character. The film shows the variables he must contend with in his role as a protector of Gotham.\\				
			\hline
		\end{tabular}	
		\label{example1}
		
	\end{table}
	
	The key to the success of response selection lies in accurately matching input messages with proper responses. The matching scores can be either individually used to rank response candidates, or used as features in a learning to rank architecture. In general, matching algorithms have to overcome semantic gaps between two objects \cite{hu2014convolutional}. In the scenario of message-response matching, the problem becomes more serious, as proper responses could contain much more information than the messages. Table 1 gives an example\footnote{It is translated from Chinese.}. The response not only answers the message, but also brings in new content (e.g., the character of Batman) into the conversation. The content represents topics to talk about the movie (e.g., ``character''). Such responses can facilitate the chatbot to engage its users, because they could arouse more discussions (e.g., discussions about ``character of Batman'') and keep the conversation going.  In practice, however, selecting such responses from others is difficult, because the extra content makes the semantic gap between messages and responses even bigger.
	
	In this paper, we study the problem of message-response matching. Particularly, we aim to improve the matching between messages and responses with rich content. Inspired by the example in Table 1, our idea is that since people bring topics into responses to enrich their content, we should match messages and responses not only by themselves, but also by their topics. Topics represent a kind of prior knowledge, and in matching, only those related to messages and responses are useful. Based on this idea, we propose a topic-aware convolutional neural tensor network (TACNTN) in order to incorporate topic information into message-response matching.
	TACNTN embeds a message, a response, and their related topic information into a vector space, and exploits all the vectors for matching by neural tensors.  The message vector and the response vector are generated by  a siamese convolutional neural network (CNN), while the topic vectors come from two topic embedding layers, one for the message and the other for the response. The two layers acquire topic words of the message and the response from a Twitter LDA model \cite{zhao2011comparing} which is pre-trained using large scale social media data outside the conversation data. The topic words of the message hint the matching model topics that could be used in the response, and the topic words of the response indicate the model if the response and the message are in the same topics.  The two layers then calculate a weight for each topic word according to the message vector and the response vector. A large weight means a topic word is relevant to the message or to the response, and the word is more useful in matching.  Finally, the two layers form topic vectors by a weighted average of the embedding of the topic words.  The final matching is conducted in message-response, message-response topic, and message topic-response, and realized by neural tensors which model the relationships between the two objects in the three pairs.  TACNTN can enjoy both the powerful matching capability of CNN with neural tensors and extra topic information provided by a state-of-the-art topic model. It extends the convolutional neural tensor network \cite{qiu2015convolutional} proposed for community question answering by the topic embedding layers for message-response matching in chatbots. With the extra topic information, responses with rich content could be boosted in ranking. We conducted empirical study on a public English data set and a human annotated Chinese data set. Evaluation results show that TACNTN can significantly outperform state-of-the-art methods for message-response matching.

	Our contributions in this paper are three-folds: 1) proposal of incorporating topic information into message-response matching. 2) proposal of a topic-aware convolutional neural tensor network for matching with topics. 3) empirical verification of the effectiveness of the proposed method on public and annotated data.
	
	\section{Related Work}
	Early work on chatbots \cite{weizenbaum1966eliza} 
	relied on handcrafted templates or heuristic rules to do response generation, which requires huge effort but can only generate limited responses. Recently, researchers begin to develop data driven approaches \cite{ritter2011data,stent2014natural}. Among the effort, retrieval based methods select a proper response by matching message-response pairs \cite{hu2014convolutional,wang2015syntax,lu2013deep}, and generation based methods employ statistical machine translation techniques \cite{ritter2011data} or the sequence to sequence framework \cite{DBLP:conf/acl/ShangLL15,serban2015building,vinyals2015neural,li2015diversity,li2016persona} to generate responses.  On top of these work, conversation history is further considered to support multi-turn conversation \cite{lowe2015ubuntu,sordoni2015neural}. In this work, we study response selection in single-turn conversation for building a retrieval based chatbot. We propose a new message-response matching method that can incorporate topic information into matching.
	
	Convolutional neural networks (CNNs) \cite{collobert2011natural} have been proven effective in many NLP tasks such as text classification \cite{kim2014convolutional}, entity disambiguation \cite{sun2015modeling}, answer selection \cite{yang2015wikiqa}, tag recommendation \cite{weston2014tagspace}, web search \cite{shen2014latent}, sentiment classification \cite{dos2014deep}, sequence prediction \cite{liconvolutional} and sentence matching \cite{hu2014convolutional}. In sentence matching, recent progress includes MultiGranCNN proposed by Yin et al. \cite{yin2015multigrancnn} who match a pair of sentences on multiple granularity, and the work of Pang et al. \shortcite{pang2016text} in which a CNN architecture in image recognition is employed for sentence matching.  In this work, we study message-response matching which is a special case of sentence matching but important for building retrieval-based chatbots. We extend the convolutional neural tensor network \cite{qiu2015convolutional,socher2013reasoning} by topic embedding layers which enable us to leverage extra topic information in matching to boost responses with rich content.

	\section{Problem Formalization}\label{probform}
	Suppose that we have a data set $\mathcal {D} = \{(y_i,m_i,r_i)\}_{i=1}^N$, where $m_i$ and $r_i$ represent an input message and a response candidate respectively, and $y_i\in \{0,1\}$ denotes a class label. $y_i=1$ means $r_i$ is a proper response for $m_i$, otherwise $y_i=0$. Each $m_i$ in $\mathcal{D}$ corresponds to a topic word set $W_{m,i}=\{w_{m,i,1}, \ldots, w_{m,i,n}\}$, and each $r_i$ in $\mathcal{D}$ has a $W_{r,i}=\{w_{r,i,1}, \ldots, w_{r,i,n}\}$ as topic words. Our goal is to learn a matching model $g(\cdot,\cdot)$ with $\mathcal{D}$ and $\{\cup_{i=1}^N W_{m,i}, \cup_{i=1}^N W_{r,i}\}$. For any message-response pair $(m,r)$, $g(m,r)$ returns a matching score which can be utilized to rank response candidates for $m$.
	
	To learn $g(\cdot, \cdot)$, we need to answer two questions: 1) how to obtain topic words, and 2) how to incorporate topic words into matching. In the following sections, we first present our method on topic word generation, then we elaborate on our matching model and learning approach.
	
	\section{Topic Word Generation}
\begin{figure}[]		
		\begin{center}
			\includegraphics[width=5cm,height=3.5cm]{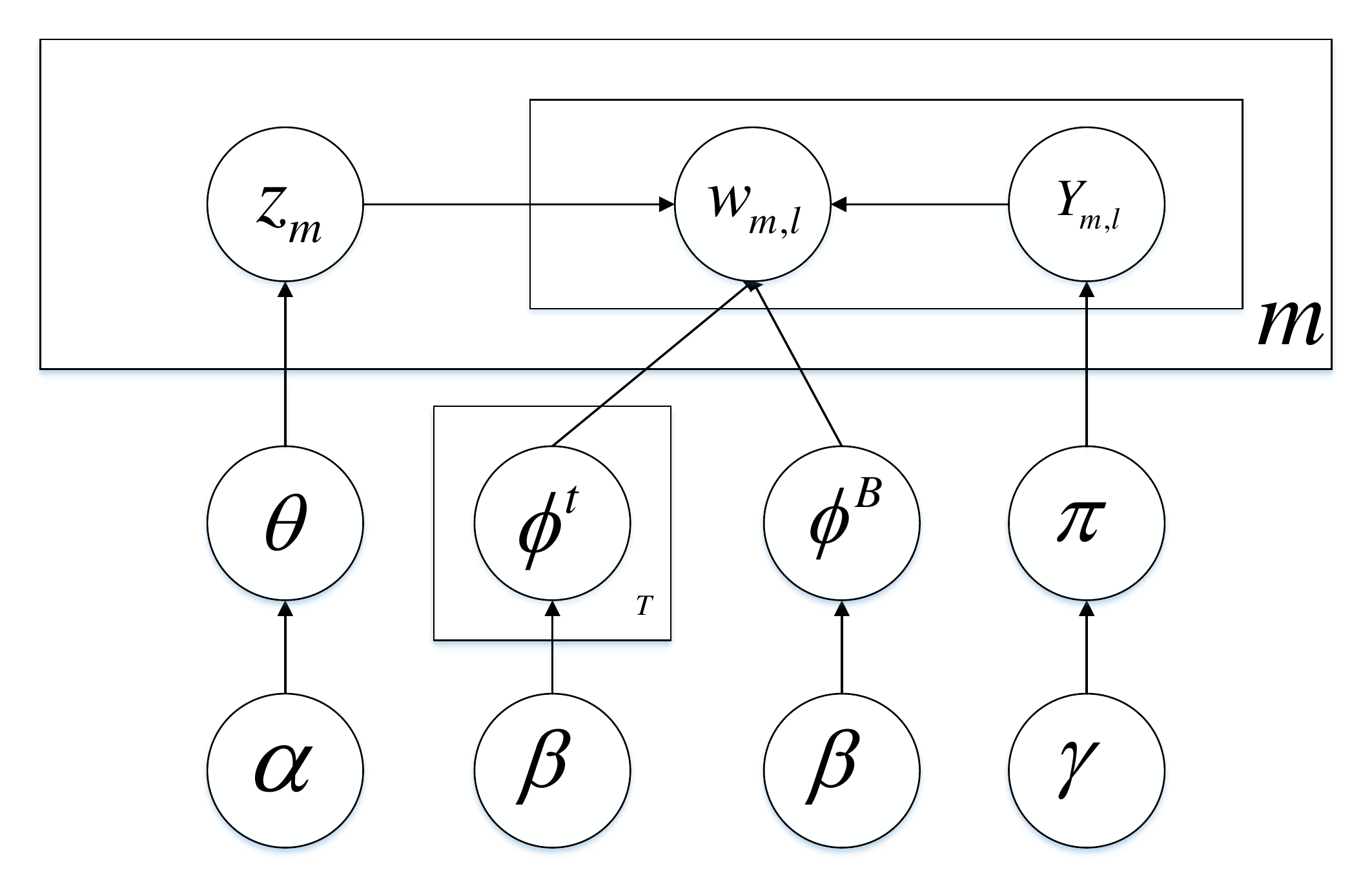}
		\end{center}
		
		\caption{Graphical model of Twitter LDA}\label{fig:lda}
	\end{figure}

	We employ a Twitter LDA model \cite{zhao2011comparing}, which is the state-of-the-art topic model for short texts, to generate topic words for messages and responses.  Twitter LDA assumes that each piece of text (a message or a response) corresponds to one topic, and each word in the text is either a background word or a topic word under the topic of the text. Figure \ref{fig:lda} shows the graphical model of Twitter LDA.
	
	We estimate the parameters of Twitter LDA using a collapsed Gibbs sampling algorithm \cite{zhao2011comparing}. After that, we use them to assign a topic to each $m_i$ and $r_i$ in $\mathcal{D}$.  To obtain the topic word sets, we define the salience of a word $w$ regarding to a topic $t$ as
	\begin{equation} \label{scoreword}
	s(w,t) = \frac{c_w^t}{c_w} \cdot c_{w}^t,
	\end{equation}
	where $c_w^t$ is the number of times that word $w$ is assigned a topic $t$ in the training data and $c_w$ is the number of times that $w$ is determined as a topic word in the training data. Equation (\ref{scoreword}) means that the salience of a word regarding to a topic is determined by the frequency of the word under the topic (i.e., $c_w^t$) and the probability of the word only belonging to the topic (i.e., $\frac{c_w^t}{c_w}$). $\frac{c_w^t}{c_w}$ plays a similar role to IDF in information retrieval, and is capable of reducing the importance of common words like "yes" and "cause" to topic $t$. With Equation (\ref{scoreword}), we select top $n$ words regarding to the topic of $m_i$ and the topic of $r_i$ to form the topic word set $W_{m,i}$ and $W_{r,i}$ respectively.
	
	In our experiments, we trained Twitter LDA models using large scale questions from Yahoo! Answers and posts from Sina Weibo. The data provides topic knowledge apart from that in message-response pairs to the learning of message-response matching. The process is similar to how people learn to respond in conversation: they become aware of what can be talked about from Internet, especially from social media, and converse with others based on what they learned.  	
	
	Note that in addition to LDA, one can employ other techniques like tag recommendation \cite{wu2016improving} or keyword extraction \cite{wu2015mining} to generate topic words. One can also get topic words from other resources like wikipedia and other web documents. We leave the discussion of these extensions as our future work. 
	

	\section{Topic-aware Convolutional Neural Tensor Network}
	\begin{figure*}[t]		
		\begin{center}
			\includegraphics[width=11cm,height=5.5cm]{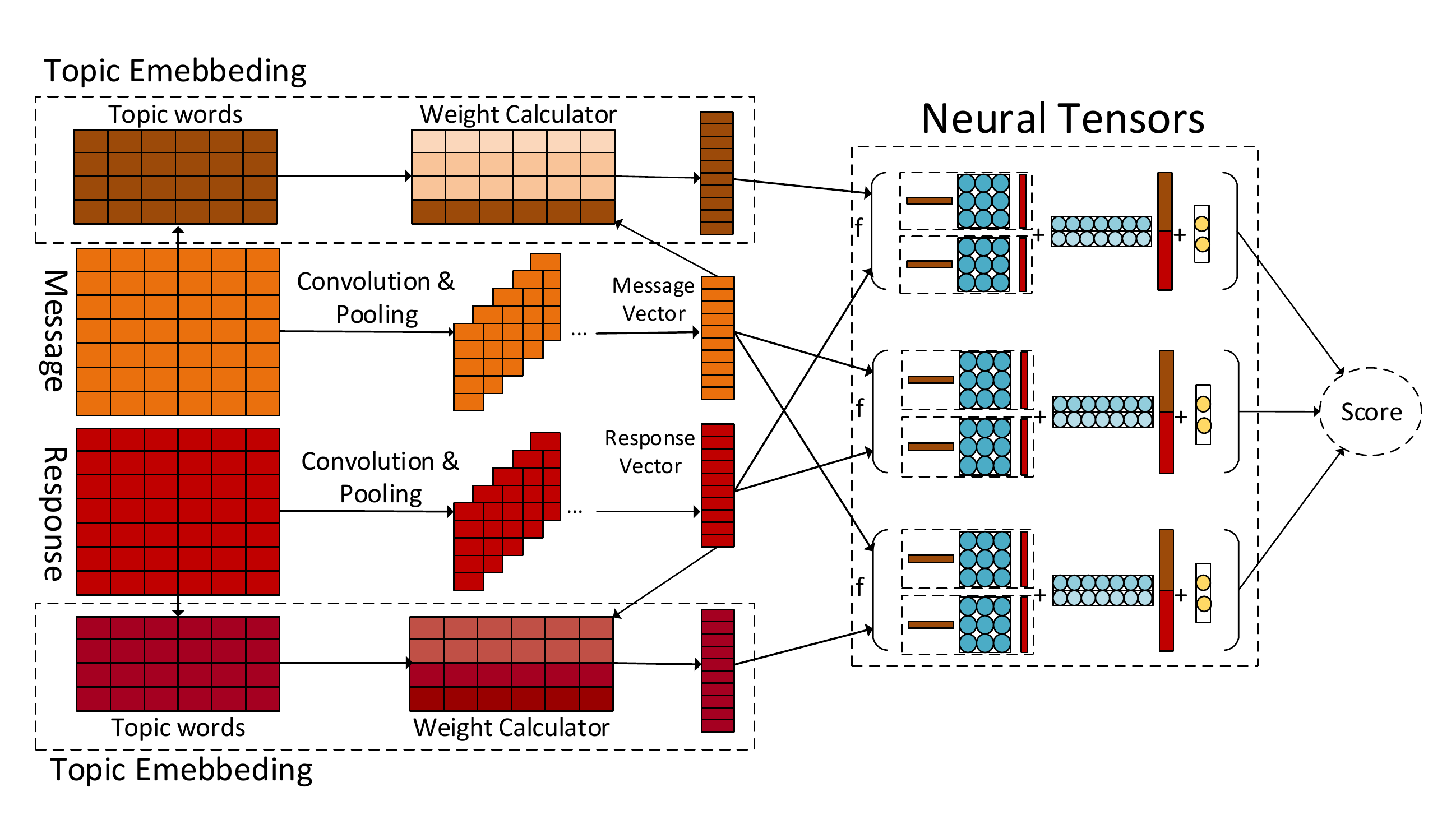}
		\end{center}
		\vspace{-6mm}
		\caption{The architecture of our model}\label{fig:framework}
	\end{figure*}	
	We propose a topic-aware convolutional neural tensor network (TACNTN) to leverage the topic words obtained from Twitter LDA in message-response matching. Figure \ref{fig:framework} gives the architecture of our model. Given a message $m$ and a response $r$, our model embeds them into a vector space by a siamese neural network that consists of two convolutional neural networks (CNNs) with shared weight. For each sentence (either $m$ or $r$), CNN first looks up a word embedding table and forms a sentence matrix $S=\left[v_1, v_2,\ldots,v_s\right]$ as input, where $v_j \in \mathbb{R}^d$ is the embedding of the $j$-th word and $s$ in the maximum length of the sentence. Note that if the length of a sentence does not reach $s$, we put all-zero padding vectors after the last word of the sentence until $s$. CNN then alternates 1D convolution operations and 1D max-pooling operations, and transforms the message matrix and the response matrix to a message vector $\vec{m}$ and a response vector $\vec{r}$ respectively. Let $z^{(l,f)} = \left[z_1^{(l,f)}, z_2^{(l,f)}, \ldots, z_{s^{(l,f)}}^{(l,f)}\right]$ denotes the output of the $l$-th layer under the $f$-th feature map (among $F_l$ of them), where $z_j^{(l,f)} \in \mathbb{R}^{d^{(l,f)}}$ and $z^{(0,f)}=S$. In convolution, CNN slides a window with width $k_1^{(l,f)}$ on $z^{(l,f)}$ and splits $z^{(l,f)}$ into several segments. For the $i$-th segment $\mathbf{z}_i^{(l,f)} = \left[z_i^{(l,f)}, \ldots, z_{i+k_1^{(l,f)}}^{(l,f)}\right]$, the output of convolution is
	\begin{equation}
	z_i^{(l+1,f)} =\sigma\left(\mathbf{z}_i^{(l,f)}\mathbf{W}^{(l,f)} + \mathbf{b}^{(l,f)}\right),
	\end{equation}
	where $\mathbf{W} ^{(l,f)} \in \mathbb{R}^{k^{(l,f)}}$ and $\mathbf{b}^{(l,f)} \in \mathbb{R}^{d^{(l,f)}}$ are parameters, and $\sigma(\cdot)$ is an activation function. In max-pooling, the output of convolution is shrunk in order to enhance robustness. Let $k_2^{(l,f)}$ denote the width of the window for max-pooling, then the output of max-pooling is
	\begin{equation}
	z_i^{(l+1,f)} = \max \left(z_i^{(l,f)} \ldots z_{i+k_2^{(l,f)}}^{(l,f)}\right),
	\end{equation}
	where $\max(\cdot)$ is an element-wise operator over vectors. CNN obtains $\vec{m}  \in \mathbb{R} ^ n$ and $\vec{r} \in \mathbb{R} ^ n$ by concatenating the vectors from the final layer.
	
	To leverage the topic words for matching, TACNTN utilizes topic embedding layers to transform the topic words to two topic vectors, one for $m$ and the other for $r$. Given $W_m=\{w_{m,1}, \ldots, w_{m,n}\}$ as the topic word set for $m$, we construct a matrix $\mathbf{T}_m = \left[e_{m,1}, \ldots, e_{m,n}\right]^\top$ by looking up a word embedding table for each word in $W_m$. We then calculate weights of the topic words by
	\begin{equation}\label{topic1}
	\omega_{m} =\mathbf{T}_m \cdot \mathbf{A} \cdot \vec{m},
	\end{equation}
	where $\mathbf{A}  \in \mathbb{R} ^ {d \times n}$ is a linear transformation learned from training data, and $\forall j$, $\omega_{m,j} \in \omega_m$ is the weight for the $j$-th word in $W_m$. We scale $\omega_{m,j}$ to $[0,1]$ by
	\begin{equation}\label{topic2}
	\alpha_{m,j}=\frac{exp\left(\omega_{m,j}\right)}{\sum_{p=1}^{n}exp\left(\omega_{m,p}\right)}.
	\end{equation}
	Finally, we form a topic vector $\vec{t}_{m}$ by a linear combination of the topic words:
	\begin{equation}\label{topic3}
	\vec{t}_{m} = \sum_{j=1}^{n} \alpha_{m,j} e_{m,j}.
	\end{equation}
	
	Following the same technique, we have a topic vector $\vec{t}_{r}$ for response $r$. From Equation (\ref{topic1}), (\ref{topic2}), and (\ref{topic3}), we can see that the more important a topic word is, the more contributions it will make to the topic vector. The importance of topic words are determined by both themselves and the message (or the response). The idea here is inspired by the attention mechanism proposed for machine translation \cite{bahdanau2014neural}. We borrow the idea of the attention mechanism here, because it well models the ineuiqvalent contribution of topic words in matching. Topic words that are more relevant to the message or to the response are more useful in matching.
	
	We calculate a matching score for $(m,r)$ by neural tensor networks (NTNs) \cite{socher2013reasoning,qiu2015convolutional}. The advantage of NTN is that it enables us to build a matching function in a bottom-up way, that is we can first model relationships between message-response, message-response topic, and message topic-response, then synthesize these sub-matching elements to a final matching score. Specifically, given $\vec{m}$ and $\vec{r}$, a neural tensor $s(\vec{m},\vec{r})$ is defined as
	\begin{equation}
	s(\vec{m},\vec{r}) = \mathbf{f}\left(  \vec{m} ^\intercal  \mathbf{M^{[1:h]}}\vec{r} + \mathbf{V} \left[ \vec{m}^\intercal, \vec{r}^\intercal \right]^\intercal + \mathbf{b} \right),
	\end{equation}
	where $f(\cdot)$ is a nonlinear function, and $\mathbf{M^{[1:h]}}  \in \mathbb{R} ^ {n \times h \times n}$ is a tensor.  The result of the bilinear tensor product $\vec{m} \mathbf{M^{[1:h]}}\vec{r} $ is a vector $\vec{v} \in \mathbb{R} ^ h$ with each entry a matching of $m$ and $r$ parameterized by a slice $k$ of $\mathbf{M}$, $k = 1, 2... h$. $\mathbf{V}  \in \mathbb{R} ^ {h \times 2n}$ and $\mathbf{b}  \in \mathbb{R} ^ h$ are the other two parameters. We employ three such neural tensors and let them individually operate on  $\left(\vec{m}, \vec{r}\right)$, $\left(\vec{m}, \vec{t}_r\right)$, and $\left(\vec{r}, \vec{t}_m\right)$, resulting in 
	three vectors $s(\vec{m},\vec{r})$, $s(\vec{m},\vec{t}_r)$, and $s(\vec{r},\vec{t}_m)$.  Each vector models the matching between the two objects from multiple perspectives parameterized by the slices of the tensor. With these vectors, we define $g(m,r)$ as
	\begin{equation}
	g(m,r) = \mathbf{h} \left( \mathbf{w}^\intercal \left[s(\vec{m},\vec{r}), s(\vec{r},\vec{t}_{m}), s(\vec{m}, \vec{t}_{r}) \right] + \mathbf{b_2} \right),
	\end{equation}
	where $\mathbf{h}$ is a softmax function, and $\mathbf{w}$ and $\mathbf{b_2}$ are parameters. We extend the convolutional neural tensor network (CNTN) proposed for community question answering \cite{qiu2015convolutional} by topic embedding layers and apply the new model to the problem of message-response matching. The model inherits the matching power from CNTN, and  naturally incorporates extra topic information into matching.      
	
	Note that in TACNTN, topic learning and matching are conducted in two steps. This is because by this means we can leverage data other than message-response pairs for matching. For example, we can estimate topic words from questions in Yahoo! Answers and use them in message-response matching. These data provides extra topic information other than that in message-response pairs. Our model explicitly utilizes such information as prior, and that is why we call it ``topic-aware''.   This is more close to how people learn to respond in conversation: before conversation, they have already had some knowledge learned from other places (e.g., social media) in their mind.

	We learn $g(\cdot, \cdot)$ by minimizing cross entropy \cite{levin1988accelerated} with $\mathcal{D}$. Let $\Theta$ denote the parameters in our model. Our objective function $\mathcal{L}(\mathcal{D};\Theta)$  is given by
	\begin{equation}\label{obj}
	\small
	-\frac{1}{N} \sum_{i=1}^{N} \left[y_i\log\left(g(m_i,r_i)\right)+(1-y_i)\log\left(1-g(m_i,r_i)\right)\right].
	\end{equation}
	We optimize the objective function using back-propagation and the parameters are updated by stochastic gradient descent with Adam algorithm \cite{kingma2014adam} controlling the learning rate. As regularization, we employ early-stopping \cite{lawrence2000overfitting} as it is enough to prevent over-fitting on large scale training data (1 million instances). We set the initial training rate and the batch size as $0.01$ and $200$ respectively.
	
	We implement TACNTN using Theano. In the implementation, we only use one convolution layer and one max-pooling layer, because we find that the performance of the model does not get better with the number of layers increased. We use ReLU \cite{dahl2013improving} as the activation function $\sigma(\cdot)$ and Tanh as the activation function $f(\cdot)$ in neural tensors. The code is shared at \url{https://github.com/MarkWuNLP/TACNTN}. 
	
	\section{Experiment}
	We tested our model on a public English data set and an in-house Chinese data set.
	\subsection{Experiment Setup}
	The English data set is the Ubuntu Corpus \cite{lowe2015ubuntu} which consists of a large number of human-human dialogues about Ubuntu-related technique support collected from Ubuntu chat rooms. Each dialogue contains at least $3$ turns, and we only kept the last turn as we focus on single-turn conversation in this work.  We used the data pre-processed by Xu et al. \cite{xu2016incorporating}\footnote{\url{https://www.dropbox.com/s/2fdn26rj6h9bpvl/ubuntu data.zip?dl=0}}, in which all urls and numbers were replaced by ``$\_url\_$'' and ``$\_number\_$'' respectively to alleviate the sparsity issue.  The training set contains $1$ million message-response pairs with a ratio $1:1$ between positive and negative responses, and both the validation set and the test set have $0.5$ million message-response pairs with a ratio $1:9$ between positive and negative responses. All the negative responses are randomly sampled rather than labeled by human annotators. 
	\begin{table*}[t] 
		\vspace{-3mm}
		\small
		\caption{Evaluation results on the Ubuntu data and the Tieba data}
		\vspace{1mm}
		\centering
		\linespread{1.1}\selectfont
		\begin{center}
			\centering
			\(
			\begin{tabular}{l|ccccc|ccc}
			\thickhline             &   \multicolumn{5}{c|}{\textbf{Ubuntu data}}    &        \multicolumn{3}{c}{\textbf{Tieba data}}        \\
			&  R$_2$@1     &  R$_5$@1    &  R$_{10}$@1 &  R$_{10}$@2&  R$_{10}$@5        &     MAP     &    MRR     &         P@1     \\ \hline
			Random         &     0.500      &     0.200      &     0.100  &0.200 &0.500    &          0.642      &     0.695      &     0.524         \\
			Cosine         &     0.681      &     0.470      &       0.383 & 0.482 &  0.686  &         0.597      &     0.662      &     0.553         \\
			Translation       &     0.721      &     0.502      &     \textbf{0.393}    & 0.507 &  0.727      &     0.710      &     0.760      &     0.658           \\
			DeepMatch$_{topic}$       &     0.593      &     0.345     &          0.248   & 0.376 &  0.693    &     0.677      &     0.725           &     0.594      \\
			MLP       &     0.651      &     0.362      &     0.256   &0.380 & 0.703  &         0.653      &     0.712      &     0.550         \\
			CNTN   &     0.743     &     0.489      &     0.349      &          0.512      &    0.797      &     0.731      &    \textbf{0.797}      &     0.670      \\	
			LSTM &     0.725     &     0.494     &    0.361    &  0.529   & 0.801 &     0.732      &     \textbf{0.797}     &     0.670          \\ 
			Arc1   &     0.665     &     0.372     &    0.221    &  0.360   & 0.684 &     0.698      &     0.771     &     0.640          \\ 
			Arc2   &     0.736     &     0.508      &    0.380   &  0.534   & 0.777   &          0.708      &     0.783      &     0.660          \\ 
			\hline
			TACNTN        & \textbf{0.759} & \textbf{0.520} & 0.382 &  \textbf{0.544}  & \textbf{0.809} &  \textbf{0.749} & \textbf{0.804} & \textbf{0.688} \\\hline
			\thickhline
			\end{tabular}
			\)
		\end{center}
		\label{exp:annotated} \vspace{-6mm}
	\end{table*}
	We built the Chinese data set from Baidu Tieba which is the largest Chinese forum allowing users to post and comment to others' posts. We first crawled $0.6$ million text post-comment pairs as positive message-response pairs (i.e., an $(m,r)$ with a $y=1$). Then, for each post, we randomly sampled another comment from the $0.6$ million data to create a negative message-response pair. The two sets together form a training set with $1.2$ million instances. Following the same procedure, we built a validation set with $50,000$ instances apart from those in training. To construct a test set, we simulated the process of a retrieval-based chatbot: we first indexed the $0.6$ million post-comment pairs by an open source Lucene.Net\footnote{\url {http://lucenenet.apache.org}}. Then, we crawled another $400$ posts that are in the training set and the validation set as test messages. For each test message, we retrieved several similar posts from the index, and collected all the responses associated with the similar posts as candidates. We recruited three human labelers to judge if a candidate is a proper response to a test message. A proper response means the response can naturally reply to the message without any contextual information. Each candidate response received three labels and the majority of the labels was taken as the final decision. After removing messages without any proper responses (i.e., two or more labels are $0$), we obtained $328$ test messages with $3,418$ responses. On average, each test message has $10.4$ labeled responses, and the ratio between positive and negative responses is 3:2. 
	
	For the English data, we crawled $8$ million questions (title and body) from the ``Computers \& Internet'' category in Yahoo! Answers, and utilized these data to train the Twitter LDA model. Word embedding tables were initialized using the public word vectors available at \url{http://nlp.stanford.edu/projects/glove} (trained on Twitter). For the Chinese data, we trained the Twitter LDA model and the word vectors for initializing embedding tables using $30$ million posts crawled from Sina Weibo.  In both data, the dimension of word vectors is $100$.

	On the Ubuntu data, we followed Lowe et al. \cite{lowe2015ubuntu} and employed recall at position $k$ in $n$ candidates ($R_n@k$) as evaluation metrics, while on the human annotated Chinese data, we employed mean average precision (MAP) \cite{baeza1999modern}, mean reciprocal rank (MRR) \cite{voorhees1999trec}, and precision at position 1 (P@1) as evaluation metrics.
	
	\subsection{Baseline}
	%
	
	We considered the following models as baselines:
	
	\textbf{Cosine:} we calculated cosine similarity between a message and a response using their tf-idf weighted vectors. 
	
	\textbf{Translation model:} we learned word-to-word translation probabilities using GIZA++\footnote{http://www.statmt.org/moses/giza/GIZA++.html} by regarding messages in training sets as a source language and their positive responses as a target language. Following \cite{ji2014information}, we used translation probability $p(response|message)$ as a matching score.
	
	\textbf{Multi-layer perceptron (MLP)}: a message and a response were represented as vectors by averaging their word vectors. The two vectors were fed to a two-layer feedforward neural network to calculate a matching score. MLP shared the embedding tables with our model. The first hidden layer has 100 nodes, and the second hidden layer has 2 nodes.
	
	\textbf{DeepMatch$_{topic}$}: the matching model proposed in \cite{lu2013deep} which only used topic information to match a message and a response.
	
	\textbf{LSTM}:  the best performing model in \cite{lowe2015ubuntu}.  A message and a response are separately fed to a LSTM network and matching score is calculated with the output vectors of the LSTM networks. 
	
	\textbf{CNNs}: the CNN models proposed by Hu et al. \shortcite{hu2014convolutional}, namely Arc1 and Arc2. The number of feature maps and the width of windows are the same as our model.
	
	\textbf{CNTN}: the convolution neural tensor network \cite{qiu2015convolutional} proposed for community question answering.

	
	\subsection{Parameter Tuning}
	We tuned parameters according to R$_2$@1 in the Ubuntu data and P@1 in the Tieba data. For Twitter LDA, we set $\alpha = 1/T$, $\beta = 0.01$, $\gamma = 0.01$ in the Dirichlet priors. We tuned the number of topics (i.e. $T$) in $\{20,50,100,200\}$ and the maximum iteration number of Gibbs sampling in $\{100,200, \ldots ,1000\}$. The best combination for both data sets is $(200,1000)$. The number of topic words was tuned in $\{10,20 , \ldots , 100 \}$ and set as $50$ finally. In CNN based models, we set the maximum sentence length (i.e., $s$) as $20$. We tuned the number of feature maps in $\{10,50,100, 200\}$ and found that $50$ is the best choice. We tuned the window size in $\{1,2,3,4\}$ and set it as $3$ for convolution and pooling layers.  We varied the number of slices in neural tensors (i.e.,$h$) in $\{1,2, \ldots 10\}$ and set it as $8$.  
	
	\begin{table}[]
		\small
		\caption{Variants of TACNTN \label{exp:varience}}	
		\centering
		\begin{tabular}{c|c|c|c|c|c}
			\thickhline
			\multicolumn{6}{c}{\textbf{Ubuntu data}}  \\ \hline
			& Exp & Avg  & Msg & Res  & Full \\ \hline
			
			R$_2$@1 & 0.750 & 0.755 & 0.754 & 0.749 & 0.759\\ \hline
			R$_5$@1 &0.499 & 0.508 & 0.498 & 0.498 & 0.520\\ \hline
			R$_{10}$@1&0.364  & 0.373 & 0.357 & 0.362 & 0.382\\ \hline
			R$_{10}$@2&0.528  & 0.538 & 0.528 & 0.538 & 0.544\\ \hline
			R$_{10}$@5&0.792  & 0.811 & 0.804 & 0.805 & 0.809\\ \hline
			\multicolumn{6}{c}{\textbf{Tieba data}}  \\ \hline
			& Exp & Avg  & Msg & Res  & Full \\ \hline
			MAP&0.734  & 0.744 & 0.744 & 0.732 & 0.749\\ \hline
			MRR&0.802  & 0.798 & 0.799 & 0.800 & 0.804\\ \hline
			P@1 &0.671 & 0.677 & 0.677 & 0.679 & 0.688\\ 
			\thickhline
		\end{tabular}
	\end{table}
	\subsection{Evaluation Results}
	Table \ref{exp:annotated} reports evaluation results on the English data and the Chinese data. We can see that on most metrics, our method performs better than baselines, and the improvement is statistically significant (t-test with $p$-value $\leq 0.01$).

	Both DeepMatch$_{topic}$ and MLP perform badly, indicating that we cannot simply rely on topics and word embeddings to match messages and responses. Among CNN based models, Arc1 is the worst. This is consistent with the conclusions drawn by the existing work. Our model consistently outperforms all baseline methods on both data sets.  The result verified the effectiveness of the topic information in message-response matching.
	
	In the Ubuntu data, all models became worse when the number of candidates (i.e., $n$ in $R_n@k$) increased and better when $k$ increased. This is because a test message only has one positive response. Obviously, ranking the only positive one to the top becomes more difficult when more negative competitors come in, but things become easier when we allow the positive one to be ranked at lower positions.

	\subsection{Discussions}
	We first examine if the topic information can improve the matching between messages and responses with rich content.  Table \ref{example2} gives a detailed analysis of the example in Table 1 which is from Tieba data. The best performing baseline CNTN failed on this case because of the high asymmetry between the message and the response on the information they contain. On the other hand, message topics like ``character'' matched the content of the response, and response topics indicated that the message and the response are in the same topics. With these signals, our model successfully overcame the semantic gap between the message and the response. We also compared our model with CNTN on short (less than $10$ words) and long (more than or equal to $10$ words) responses. Table \ref{Acc on Length} reports classification accuracy on the two data sets. We use classification accuracy because we make comparison in terms of different responses. We can see that almost all improvement of our model came from long responses. Since long responses contain rich content, the results provided quantitative evidence to our claim that we can improve the matching between messages and responses with rich content using topic information. 
	
	
	Finally, we compared TACNTN with several variants including TACNTN$_{exp}$ in which we expand the message and the response with their topic words and conduct matching using CNTN on the expanded texts, TACNTN$_{avg}$ in which the topic vectors are not learned with the message vector and the response vector but average of the embedding of the topic words, TACNTN$_{msg}$ in which only message topics are considered, and TACNTN$_{res}$ in which only response topics are considered. The four variants are denoted as Exp, Avg, Msg and Res respectively. Table \ref{exp:varience} shows that 1) utilizing topic information in a straightforward way (i.e., expansion) does not help too much; 2) learning topic vectors with the message vector and the response vector is helpful to matching as they may help filter noise in the topic words, and that is why TACNTN outperforms TACNTN$_{avg}$; 3) both the message topic vector and the response topic vector are useful, and we cannot leave either of them out.  
	

	%

	\begin{table}[]
		\small
		\caption{Analysis of TACNTN}	
		\centering
		\subtable[A case study]{
			\label{example2}
			\begin{tabular}{m{2.5cm}|m{5cm}}
				\hline
				Message & Is the new Batman movie worth watching?  \\ \hline
				Message topic words and weights & movie: 0.198 , character: 0.187, plot :0.087 \\ \hline
				Response & I swear you won't regret watching it. We finally get Batman as a fully rendered character. The film shows the variables he must contend with in his role as a protector of Gotham.	\\ \hline
				Response topic words and weights & movie: 0.405, character: 0.217 , hero: 0.184 \\ \hline
				CNTN & 0.447 \\ \hline
				TACNTN & 0.959 \\ \hline
				Label & Relevant \\ \hline
			\end{tabular}
		}
		\subtable[Quantitative analysis]{
			\label{Acc on Length}
			\begin{tabular}{m{2.5cm}|m{2.5cm}|m{2cm}}
				\hline
				Response Length & $ < $10 & $ \geq$ 10 \\ \hline
				CNTN on Ubuntu & 0.673 & 0.702\\ \hline
				TACNTN  on Ubuntu & 0.675 & 0.723\\ \hline
				CNTN on Tieba & 0.586 & 0.589\\ \hline
				TACNTN on Tieba &0.588 & 0.612\\ \hline
			\end{tabular}
		}
		\vspace{-4mm}	
	\end{table}
	
	\section{Conclusion}
	This paper proposed a TACNTN model that can leverage topic information in message-response matching for retrieval-based chatbots. Experimental results show that our model can significantly outperform state-of-the-art matching models on public and human annotated data sets.
	
	\small
	\bibliographystyle {aaai16}
	\bibliography {aaai}
	
\end{document}